\documentclass[conference]{IEEEtran}
\usepackage{cite}
\usepackage{amsmath,amssymb,amsfonts}
\usepackage{algorithmic}
\usepackage{graphicx}
\usepackage{booktabs}
\usepackage{textcomp}
\usepackage{xcolor}
\usepackage{siunitx}
\usepackage{url}
\usepackage{mathtools}
\def\BibTeX{{\rm B\kern-.05em{\sc i\kern-.025em b}\kern-.08em
    T\kern-.1667em\lower.7ex\hbox{E}\kern-.125emX}}

\begin{document}

\graphicspath{{images/}}
\bibliographystyle{literature/IEEEtran}
\title{FlexKalmanNet: A Modular AI-Enhanced Kalman Filter Framework Applied to Spacecraft Motion Estimation}
\IEEEspecialpapernotice{“This work has been submitted to the IEEE for possible publication. Copyright may be transferred without notice, after which this version may no longer be accessible.”}
\author{
\IEEEauthorblockN{1\textsuperscript{st} Moritz D. Pinheiro-Torres Vogt}
\IEEEauthorblockA{\textit{Chair of Space Technology} \\
\textit{TU Berlin}\\
Berlin, Germany \\
0000-0002-1409-7240}
\and
\IEEEauthorblockN{2\textsuperscript{nd} Markus Huwald}
\IEEEauthorblockA{\textit{Chair of Space Technology} \\
\textit{TU Berlin}\\
Berlin, Germany \\
0009-0009-3249-4627}
\and
\IEEEauthorblockN{3\textsuperscript{rd} M. Khalil Ben-Larbi}
\IEEEauthorblockA{\textit{Chair of Space Technology} \\
\textit{TU Berlin}\\
Berlin, Germany \\
0000-0002-8390-9302}
\and
\IEEEauthorblockN{4\textsuperscript{th} Enrico Stoll}
\IEEEauthorblockA{\textit{Chair of Space Technology} \\
\textit{TU Berlin}\\
Berlin, Germany \\
0000-0002-4760-3445}
}

\maketitle

\begin{abstract}
The estimation of relative motion between spacecraft increasingly relies on feature-matching computer vision, which feeds data into a recursive filtering algorithm. 
Kalman filters, although efficient in noise compensation, demand extensive tuning of system and noise models. 
This paper introduces FlexKalmanNet, a novel modular framework that bridges this gap by integrating a deep fully connected neural network with Kalman filter-based motion estimation algorithms. 
FlexKalmanNet's core innovation is its ability to learn any Kalman filter parameter directly from measurement data, coupled with the flexibility to utilize various Kalman filter variants. 
This is achieved through a notable design decision to outsource the sequential computation from the neural network to the Kalman filter variant, enabling a purely feedforward neural network architecture. 
This architecture, proficient at handling complex, nonlinear features without the dependency on recurrent network modules, captures global data patterns more effectively. 
Empirical evaluation using data from NASA's Astrobee simulation environment focuses on learning unknown parameters of an Extended Kalman filter for spacecraft pose and twist estimation. 
The results demonstrate FlexKalmanNet's rapid training convergence, high accuracy, and superior performance against manually tuned Extended Kalman filters.
\end{abstract}

\begin{IEEEkeywords}
Kalman filter, spacecraft motion estimation, artificial neural network
\end{IEEEkeywords}
\section{Introduction}\label{sec:introduction} 

Recent advancements in vision sensor technology have significantly impacted on-board pose estimation and tracking capabilities of space objects, a key technology for On-Orbit Servicing (OOS) and Active Debris Removal (ADR) missions. 
These missions require real-time information about the target's pose relative to the servicer spacecraft to safely and efficiently execute autonomous rendezvous and docking trajectories. 
The main challenge is the limited prior knowledge about the target object's structure and motion. 
Vision sensors acquire two-dimensional images of the three-dimensional scene. The 3D to 2D mapping can be modeled in several ways, but all of them lead to nonlinear measurement equations.
This paper proposes a filtering framework for handling the complications encountered in such scenarios. 

While methods like the Kalman Filter (KF) offer noise compensation with lower computational demands than alternatives like the Particle Filter (PF), they require precise knowledge or extensive tuning of inherently nonlinear systems and noise models. 
The integration of Artificial Intelligence (AI) with KFs presents an opportunity to learn these statistics and to automate tuning, potentially surpassing manual calibration in efficiency. 
In their comprehensive survey, Kim et al.~\cite{kim2022review} categorized various AI-based KF approaches, highlighting their effectiveness in dynamic environments. 
For example, Jiang and Nong's~\cite{jiang2020trainable} approach treats KFs as a kernel in Recurrent Neural Networks (RNNs) to dynamically predict key parameters, improving adaptability and accuracy, particularly in estimating noise covariance matrices. 
With a similar objective, Ullah et al.~\cite{ullah2019improving} developed an Artificial Neural Network (ANN)-based module to dynamically update the measurement noise covariance matrix $\boldsymbol{R}$ in response to environmental changes, yielding more accurate results than conventional KFs. 
Expanding on this, Jouaber et al.~\cite{jouaber2021nnakf} introduced a RNN-based KF focusing on the dynamic process noise covariance matrix $\boldsymbol{Q}$, a critical factor during maneuvers with increased uncertainties. 
Applying reinforcement learning, Xiong et al.~\cite{xiong2021q} combination of an Extended Kalman Filter (EKF) with Q-learning for adaptive covariance significantly improved estimation performance over traditional EKFs. 
For non-adaptive scenarios, Xu and Niu’s~\cite{xu2021ekfnet} data-driven AI approach, \textit{EKFNet}, learns optimal noise covariance values from multiple KF components, surpassing the capabilities of manually tuned EKFs. 
Similarly, Revach et al.~\cite{revach2022kalmannet} developed \textit{KalmanNet}, replacing key portions of the KF with an ANN, handling nonlinearities and model mismatches and outperforming classical KFs.

For spacecraft motion estimation, most existing algorithms overlook the twist component. 
Notable exceptions include Park and D'Amico’s work~\cite{park2023adaptive}, which integrates Convolutional Neural Network (CNN)-based pose estimation with an Unscented Kalman Filter (UKF) for both pose and twist estimation. 
However, this approach does not leverage AI for parameterization and requires substantial computational resources to adaptively tune the noise covariance matrices online. 

Altogether, current algorithms predominantly focus on single aspects of the KF's application, either in specific scenarios or for particular variants of the filter. 
This specialization limits the scope of existing solutions, leaving a gap for a more versatile and comprehensive approach. 
Addressing this gap, this work introduces FlexKalmanNet, a novel framework that combines a Deep Fully Connected Neural Network (DFCNN) with a motion estimation algorithm, specifically the Kalman filter.
FlexKalmanNet's primary contribution lies in its ability to learn any KF parameter directly from measurement data, while allowing the use of different KF variants. 
A critical innovation in this framework is the outsourcing of the recursive computations from the neural network to the KF. 
This design choice enables the use of a purely feedforward neural network architecture, adept at mapping complex and nonlinear features without relying on recurrent network modules. 
While recurrent architectures are limited to sequential patterns, the proposed architecture is capable of capturing global patterns.

The paper is organized as follows: Section 2 details the relevant EKF equations and defines the required dynamics of spacecraft relative motion, selected for the exemplary implementation. 
Following this, section 3 comprehensively outlines the FlexKalmanNet architecture. 
Section 4 covers the materials and methods utilized on the study, while Section 5 summarizes the simulation results and discusses performance aspects. 
The paper concludes with Section 6, summarizing the key lessons learned from the research and discussing limitations, ongoing tasks, and future challenges.

\section{Extended Kalman Filter for Tumbling Spacecraft State Estimation}
\label{sec:EKF}
Extensions to the KF like the EKF and UKF have been applied for state estimation in case of nonlinear systems~\cite{crassidis2011optimal}. 
The EKF, preferred for its computational efficiency, is chosen for this study. In all KF variants, state estimates are iteratively computed through two primary steps: propagation and update~\cite{crassidis2011optimal}. 
Initializing the EKF requires initial guesses for the state vector $\boldsymbol{x}_0$ and error covariance matrix $\boldsymbol{P}_0$. 
The EKF's propagation step, formulated in \eqref{eq:ekf_prediction_1} and \eqref{eq:ekf_prediction_2}, involves a one-step-ahead forecast for the current state estimate $\hat{\boldsymbol{x}}^+_{k}$ and state covariance matrix $\boldsymbol{P}^+_{k}$ following an update. 

\begin{equation}
\hat{\boldsymbol{x}}^-_{k+1}=f(\hat{\boldsymbol{x}}^+_{k}, \boldsymbol{u}_k)
\label{eq:ekf_prediction_1}
\end{equation}
\begin{equation}
\boldsymbol{P}_{k+1}^-= \boldsymbol{F}(\hat{\boldsymbol{x}}^+_{k}, \boldsymbol{u}_k) \boldsymbol{P}^+_{k} \boldsymbol{F}^\mathrm{T}(\hat{\boldsymbol{x}}^+_{k}, \boldsymbol{u}_k) + \boldsymbol{Q}_k
\label{eq:ekf_prediction_2}
\end{equation}

\noindent In these equations $f$ represents the linearized state transition model, $\boldsymbol{F}$ is the Jacobian of $f$, ${u}_k$ the control input, and $\boldsymbol{Q}_k$ is the process noise covariance matrix. 
The subscripts $k$ and $k+1$ denote the current and forecasted time steps, while the superscripts $-$ and $+$ indicate state estimates before and after a measurement update, respectively. 
Prior to the update, the Kalman gain $\boldsymbol{K}_k$ is computed using \eqref{eq:ekf_correction_3}.
Subsequently, the measurement update, elaborated in \eqref{eq:ekf_correction_4} and \eqref{eq:ekf_correction_5}, is performed. 
This step involves updating the propagations using the current measurement $\boldsymbol{\tilde{y}}_k$, the linearized observation model $h$, and its Jacobian $\boldsymbol{H}_k$, along with the measurement noise covariance matrix $\boldsymbol{R}_k$, and the state covariance matrix $\boldsymbol{P}^-_k$, where $\boldsymbol{\mathrm{I}}$ is the identity matrix.
%
\begin{equation}
\boldsymbol{K}_k= \boldsymbol{P}^-_k\boldsymbol{H}_k^\mathrm{T}(\hat{\boldsymbol{x}}^-_k)\left[\boldsymbol{H}_k(\hat{\boldsymbol{x}}^-_k) \boldsymbol{P}_k^-\boldsymbol{H}_k^\mathrm{T}(\hat{\boldsymbol{x}}^-_k) + \boldsymbol{R}_k\right]^{-1} 
\label{eq:ekf_correction_3}
\end{equation}
\begin{equation}
\hat{\boldsymbol{x}}^+_k= \hat{\boldsymbol{x}}^-_k+ \boldsymbol{K}_k\left[\tilde{\boldsymbol{y}}_k - h(\hat{\boldsymbol{x}}^-_k) \right]
\label{eq:ekf_correction_4}
\end{equation}
\begin{equation}
\boldsymbol{P}^+_k= \left[\boldsymbol{\mathrm{I}} - \boldsymbol{K}_k\boldsymbol{H}_k(\hat{\boldsymbol{x}}^-_k)\right]\boldsymbol{P}^-_k
\label{eq:ekf_correction_5}
\end{equation} 

For a servicer spacecraft closely observing a free-tumbling target in Earth orbit, a 13-component state vector is adopted, comprising orientation as a quaternion $\boldsymbol{q}$ in the Hamilton \textit{wxyz} convention, position $\boldsymbol{r}$, translational velocity $\boldsymbol{v}$, and angular velocity $\boldsymbol{\omega}$. 
\begin{equation}
\begin{split}
\boldsymbol{x} &= {[\boldsymbol{q}, \boldsymbol{r}, \boldsymbol\omega, \boldsymbol{v}]}^\mathrm{T} \\
&= {[q_{w}, q_{x}, q_{y}, q_{z}, r_x, r_y, r_z, \omega_x, \omega_y, \omega_z, v_x, v_y, v_z]}^\mathrm{T}
\end{split}
\label{eq:ekf_state_vector}
\end{equation}
Considering the brief observation time (few minutes) and proximity to the target (few meters) compared to the orbital cycle, the translational dynamics of the spacecraft are effectively modeled using a double integrator, setting aside relative orbital dynamics. 
Furthermore, for ease of modeling, we assume isotropic, torque-free rotational motion. The time-discrete state transition model $f(\boldsymbol{x})$, detailed in \eqref{eq:motion_position} to \eqref{eq:motion_angular_velocity}, along with its Jacobian $\boldsymbol{F}(\boldsymbol{x})$, are derived based on these assumptions. 
\begin{equation}
\boldsymbol{r}_{k+1} = \boldsymbol{r}_{k} + \boldsymbol{v}_{k} \Delta t
\label{eq:motion_position}
\end{equation}
\begin{equation}
\boldsymbol{v}_{k+1} = \boldsymbol{v}_{k}
\label{eq:motion_translational_velocity}
\end{equation}
\begin{equation}
\boldsymbol{q}_{k+1} = \boldsymbol{q}_{k} + \frac{\Delta t}{2} \boldsymbol\Theta_{{\mathrm{W}}_{k}} \boldsymbol{\omega}_{k}
\label{eq:motion_quaternion}
\end{equation}
\begin{equation}
\boldsymbol\omega_{k+1} = \boldsymbol\omega_{k}
\label{eq:motion_angular_velocity}
\end{equation}
The term $\Delta t$ denotes the time interval between two consecutive discrete-time steps, while $\boldsymbol\Theta_{\mathrm{W}}$ is a matrix that maps the quaternion components to the angular velocity components in the inertial world frame defined as:
\begin{equation}
\boldsymbol\Theta_{\mathrm{W}} = 
\begin{bmatrix*}[r]
-q_{x} & -q_{y} & -q_{z} \\
 q_{w} &  q_{z} & -q_{y} \\
-q_{z} &  q_{w} &  q_{x} \\
 q_{y} & -q_{x} &  q_{w}
\end{bmatrix*}
\label{eq:quaternion_theta}
\end{equation}

Further, orientations and positions are assumed to be provided by a vision-based relative navigation system. 
This allows for the derivation of the observation model $h(\boldsymbol{x})$ and its Jacobian $\boldsymbol{H}(\boldsymbol{x})$. 
Assuming the state features' noises are uncorrelated, the process noise covariance $\boldsymbol{Q}$ is modeled as a diagonal matrix~\cite{formentin2014insight}: 
\begin{equation}
\boldsymbol{Q} = \boldsymbol{\mathrm{I}}_{13} \, \boldsymbol\sigma_{Q,13}^{2}
\label{eq:ekf_process_noise}
\end{equation}
Here, $\boldsymbol\sigma_{Q,13}$ represents standard deviations for each state vector component in $\boldsymbol{x}$. Similarly, the measurement noise covariance $\boldsymbol{R}$ is defined: 
\begin{equation}
\boldsymbol{R} = \boldsymbol{\mathrm{I}}_7 \, \boldsymbol\sigma_{R,7}^{2}
\label{eq:ekf_measurement_noise}
\end{equation}
with $\boldsymbol\sigma_{R,7}$ indicating standard deviations for each measurement vector component in $\tilde{\boldsymbol{y}}$. 
The architecture of FlexKalmanNet, which enables the estimation of $\boldsymbol{Q}$ and $\boldsymbol{R}$, will be introduced in the next section.

\section{FlexKalmanNet Architecture}

FlexKalmanNet is introduced to address the evolving computational needs and restrictions, which may favor to use a linear KF or a UKF over an EKF in future scenarios. 
Offering modular treatment of the KF, it uniquely supports the interchangeable use of KF variants, a feature not seen in current approaches. 
This framework enables training a DFCNN to learn any unknown KF parameter from measurement data. 
Hence, the parameters to be learned can be chosen with full flexibility. FlexKalmanNet augments the existing state-of-the-art approaches by providing a generalized versatile framework. 
\begin{figure}[!tb]
\centering
\includegraphics[width=0.485\textwidth]{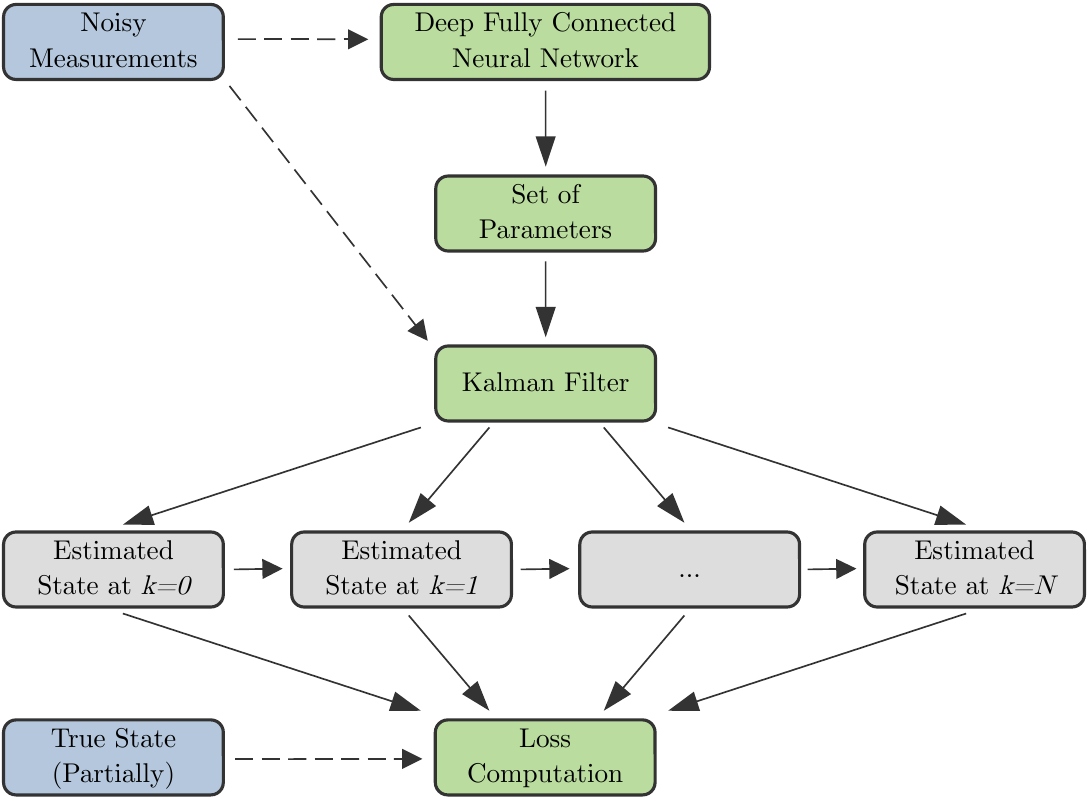}
\caption{Forward pass of the computational graph of FlexKalmanNet. The flow of activations during a forward pass is illustrated via the solid arrows. Dashed arrows represent data flows that are relevant for the framework, but not part of the computational graph. The backward pass for the gradients goes in the opposite direction of the illustrated solid lines.}
\label{fig:computational_graph}
\end{figure}

In this work, FlexKalmanNet integrates an EKF to identify the noise covariances $\boldsymbol{Q}$ and $\boldsymbol{R}$. 
The computational graph of FlexKalmanNet, detailed in Fig.~\ref{fig:computational_graph}, depicts how noisy inputs are concurrently processed by the DFCNN for global parameter learning and by the KF variant for sequential state estimation. 
Initially, the DFCNN learns from the noisy inputs to generate parameters (e.g. noise covariance) for the KF, which then estimates the system's state at sequential time steps. 
Discrepancies between estimates and available data from the true states across all time steps are quantified as loss, which is used to train the DFCNN through backpropagation (backward pass).

Unlike KalmanNet or EKFNet, FlexKalmanNet's training bypasses the limitations of sequential data learning, such as computational demands and gradient instabilities, by outsourcing the sequential computation to the KF variant. Moreover, in contrast to KalmanNet, FlexKalmanNet learns required parameters for the KF instead of providing a trained model to substitute a specific component within the KF.
This approach results in a learning model that captures global patterns effectively, leveraging the strength of DFCNNs to connect any input to any output, all without focusing on sequential patterns, as is common with RNNs.

The learning process in FlexKalmanNet follows a chained structure, where measurements fed through the network yield parameters for the KF instance to process, which eventually yields estimates utilized for the loss computation.
This structure not only has the potential to improve data processing efficiency but also provides adaptability to learn different parameters and use different KF variants. 
This shift from an end-to-end approach to a chained process represents a contribution towards a more scalable and potentially robust framework for dynamic state estimation.

\section{Materials and Methods}
We evaluated FlexKalmanNet in a space scenario with the state estimation of a free flying Astrobee robot~\cite{bualat2018astrobee} rotating at constant angular velocity. 
FlexKalmanNet learned the noise covariance parameters of the EKF from noisy measurement data and ground truth using the state transition model in Sec.~\ref{sec:EKF}.

\subsection{Data Generation}
To generate the necessary measurement data, we introduced noise to the ground truth, resulting in a dataset that realistically represented Astrobee's pose in the world frame. 
This data generation process was accomplished using the Robot Operating System (ROS)-based Astrobee Robot Software (ARS)~\cite{fluckiger2018astrobee}, which offers a simulation environment for controlling and visualizing Astrobees within a virtual International Space Station (ISS). 
To facilitate systematic testing, the constant rotation of Astrobee in the simulation is achieved through forced motion using thrusters, rather than free tumbling motion. 
However, this still effectively simulates isotropic torque-free motion. We created two datasets under different conditions:

\begin{itemize}
    \item \textbf{Dataset 1 (DS1)}: Realistic angular velocity with $\boldsymbol\omega = [0.02, 0.04, 0.06]^\mathrm{T}\,\si{\radian}/\si{\second}$, approximately twice the rotation magnitude of Earth's orbiting satellite, Envisat~\cite{silha2016comparison}.
    \item \textbf{Dataset 2 (DS2)}: Excessive angular velocity with $\boldsymbol\omega = [0.10, 0.20, 0.30]^\mathrm{T}\,\si{\radian}/\si{\second}$, used to assess the stability and sensitivity of the trained EKF.
\end{itemize}

Both datasets contain position and orientation data. 
While Orientation data, provided as unit quaternions, did not require normalization, the position data were normalized for better training convergence~\cite{huang2023normalization}. 
This normalization followed the standard score approach~\cite{patro2015normalization}, which accounts for statistical outliers typically introduced by Gaussian noise, the same type of noise added to simulate measurement noise ($\sigma_R = 0.1$, $\mu = 0$) in this study. 
For position data, the noise is directly added to the measurements. 
For quaternions, we introduce the noise as an error rotation quaternion, with each axis experiencing a rotation angle equal to the standard deviation in radians. 
This results in a probabilistic rotation error magnitude of around $\SI{10}{\degree}$. 

In the context of spacecraft stand-off observation from a safe hold point, the translational velocity was set to $\SI{0}{\meter/\second}$ during data recordings. 
Both datasets were sampled at a rate of $\SI{10}{\hertz}$ providing $16000$ samples each. 
The data were divided into training $(\SI{80}{\percent})$, validation $(\SI{10}{\percent})$, and testing $(\SI{10}{\percent})$ sets. 

\subsection{Training and Hyperparameters}
During DFCNN training, FlexKalmanNet faces the challenge of learning the parameters for the process noise covariance matrix $\boldsymbol{Q}$ and the measurement noise covariance matrix $\boldsymbol{R}$, which add up to a total of $20$ parameters to be learned. 
To compute the loss, all estimated state features from the EKF are compared to the noise-free ground truth. 
Furthermore, a loss cut is introduced to enhance learning. 
This cut excludes the initial portion of estimated states and their corresponding ground truths from the loss calculation, reducing the impact of large errors during the EKF's settling time. 
While it is possible to apply a weighting policy to all losses, the cut is chosen for loss optimization to avoid introducing bias to the learning process~\cite{byrd2019effect}.

\begin{figure*}[!tb]
\centering
\includegraphics[width=0.7\textwidth]{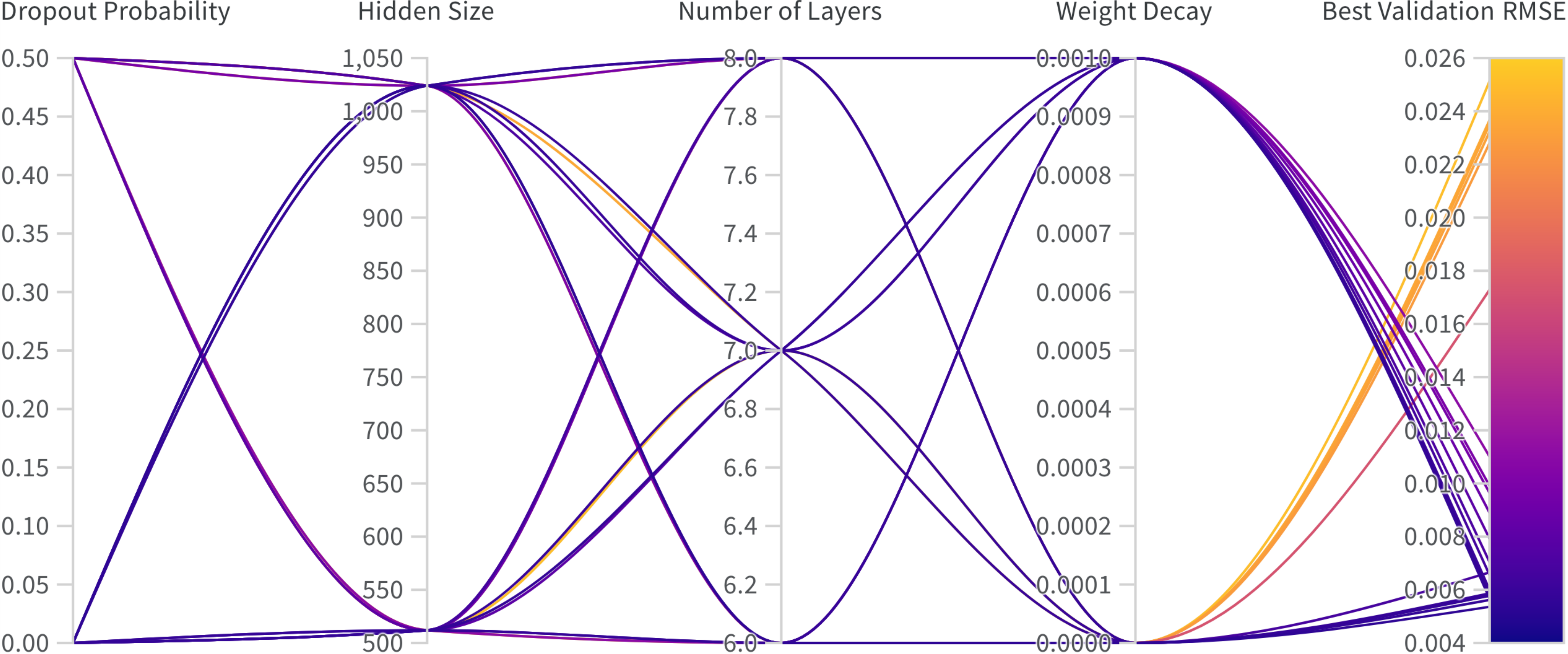}
\caption{Hyperparameter grid search mapped to the best validation loss for DS1. Each column represents a variable in the hyperparameter space that has more than one available value. Sweeps are traversing these columns from the left hand side to the right hand side, mapping their hyperparameters to their best validation loss as root mean squared error. The color map indicates the magnitude of the loss relative to the minimum and maximum loss values of all sweeps. Plot generated via Weights and Biases~\cite{wandb}.}
\label{fig:sweeps_ds1_filter_0}
\end{figure*}

Besides the loss cut, there are several hyperparameters that must be set before training. 
To determine the best hyperparameters for both datasets, a two-step approach was used: first, a random search to narrow down a reduced hyperparameter space, followed by a grid search. 
After defining the hyperparameter space, DFCNN training sessions were executed, and their performance, specifically their lowest validation losses, was recorded. Analyzing the losses across the various hyperparameters used in each run allowed for the identification of the most suitable hyperparameters (cf. Fig.~\ref{fig:sweeps_ds1_filter_0}). 
This search aimed not only to find the best hyperparameters for each dataset but also to discover a set of hyperparameters that can generalize well across datasets with varying angular velocities in similar training scenarios.  
Table~\ref{tab:hyperparameter_set} presents the best hyperparameters found for training FlexKalmanNet with the data in this study.

\begin{table}[!tb]
\centering
\caption{Best found common hyperparameter set to train FlexKalmanNet with the data of this work}
\begin{tabular}{l|l}
\hline
\hline
\addlinespace
Hyperparameter & Optimal value \\
\addlinespace
\toprule
\addlinespace
Random seed & $0$ \\
Input size & $7$ \\
Output size & $20$ \\
Batch size & $512$ \\
Hidden size & $512$ \\
Number of layers & $7$ \\
Learning rate & $1e^{-5}$ \\
Dropout probability & $0$ \\
Weight decay penalty term & $0$ \\
Loss function & MSE \\
Loss scheduler & ReduceLROnPlateau \\
Loss cut & $100$ \\
Optimizer & Adam \\
Activation function & LeakyReLu \\
Number of epochs & $100$ \\
Early stopping after epochs & $20$ \\
\addlinespace
\hline
\hline
\end{tabular}
\label{tab:hyperparameter_set}
\end{table} 

The training was conducted on a virtual server, equipped with a Solid State Drive (SSD), with $\SI{8}{\giga\byte}$ of memory, and with $4$ virtual cores accessing a \textit{AMD EPYC 7452 Rome} Central Processing Unit (CPU) providing $32$ cores with each $\SI{2.35}{\giga\hertz}$ clock speed. 
The virtual server ran on Ubuntu 20.04 and with no access to a Graphics Processing Unit (GPU). 

\subsection{Evaluation Criteria}
To evaluate the EKF using the learned parameters, we applied it to large sequences of the recorded datasets rather than just short batches. 
This approach was selected because running the EKF on short batch lengths during training may not reveal instability over time. 
Further, in real-life scenarios, measurements may only be available at a low frequency. 
Therefore, we simulated sparse measurements, making them available every $10$ discrete-time steps. 
With the ground truth available, EKF state estimates can be directly compared, using the Root Mean Squared Error (RMSE) as a measure of accuracy. 
The RMSE computes the loss between estimated and true states, preserving the original loss variable domain. 
During training, three distinct types of losses are considered. 
Firstly, the training loss evaluates how well the predictions align with the ground truth during the training phase. 
Secondly, the validation loss assesses the model's performance on data not encountered during training, offering insights into its generalization capability. 
Lastly, the test loss measures the model's performance using a separate dataset that was not part of the training process, providing an overall indicator of predictive accuracy. 

The performance of the trained EKF was eventually compared to a baseline, a manually tuned EKF. 
In the manually tuned EKF, we matched the entries of $\boldsymbol{R}$ to the ground truth noise standard deviation $(\sigma_R = 0.1)$. 
For the quaternion components and angular velocities, which involve linearized approximations, we adjusted the entries of $\boldsymbol{Q}$ to $\sigma_{Q_{\mathrm{rot}}} = 0.005$. 
However, for positions and translational velocities, where perfect model dynamics are assumed, we opted for lower values in $\boldsymbol{Q}$, specifically $\sigma_{Q_{\mathrm{trans}}} = 0.0001$.

\section{Simulation Results}
In this section, the analysis focuses on the training and performance of FlexKalmanNet. 
The goal is to evaluate the learned EKF parameters for accuracy, stability, robustness, sensitivity, and settling time. 
Additionally, a comparison is made with a manually tuned EKF, which serves as the baseline.

\subsection{Training and Parameter Convergence}
\begin{figure}[!tb]
\centering
\includegraphics[width=0.485\textwidth]{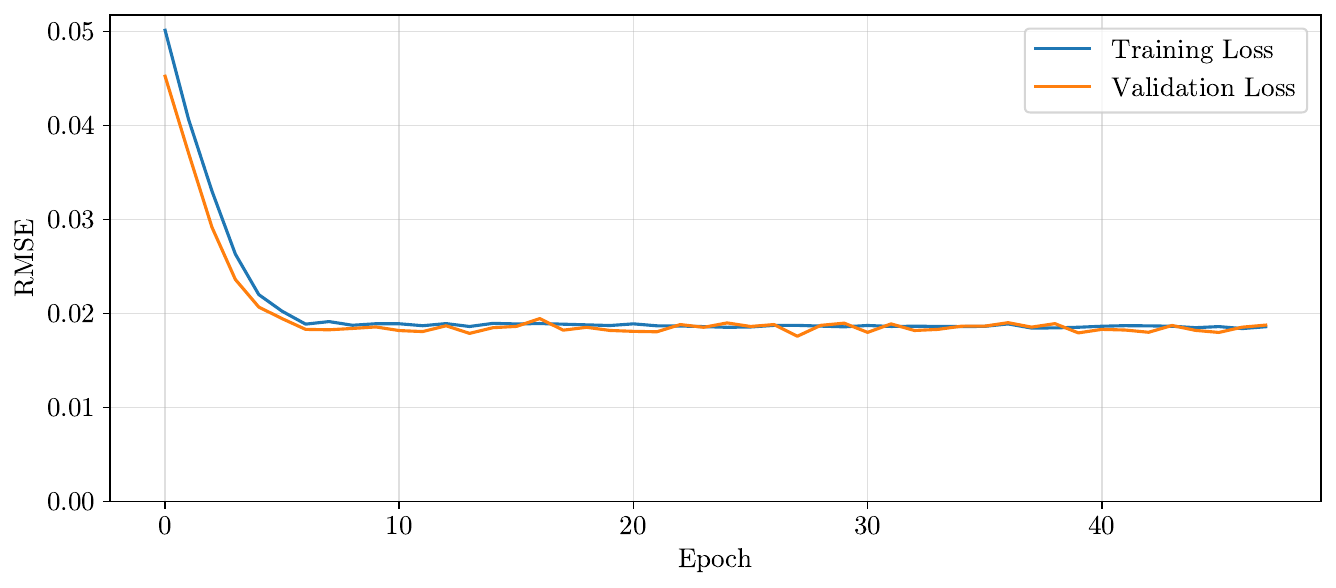}
\caption{Convergence behavior of the RMSE losses during training with DS2. Two RMSE losses of each epoch during the training of the DFCNN model using DS2 are presented: The loss from the training phase, blue, and the loss of the validation phase, orange.}
\label{fig:training_losses_ds2}
\end{figure}

In Fig.~\ref{fig:training_losses_ds2}, we can observe the convergence of FlexKalmanNet's training using DS2. 
The training appears stable and converges quickly, with convergence beginning around the sixth epoch. 
However, at epoch $27$, a negative peak in the validation loss is noticeable. 
After $20$ more epochs, the training is terminated due to early stopping because no lower loss than the one at epoch $27$ is achieved.

\begin{figure}[!tb]
\centering
\includegraphics[width=0.485\textwidth]{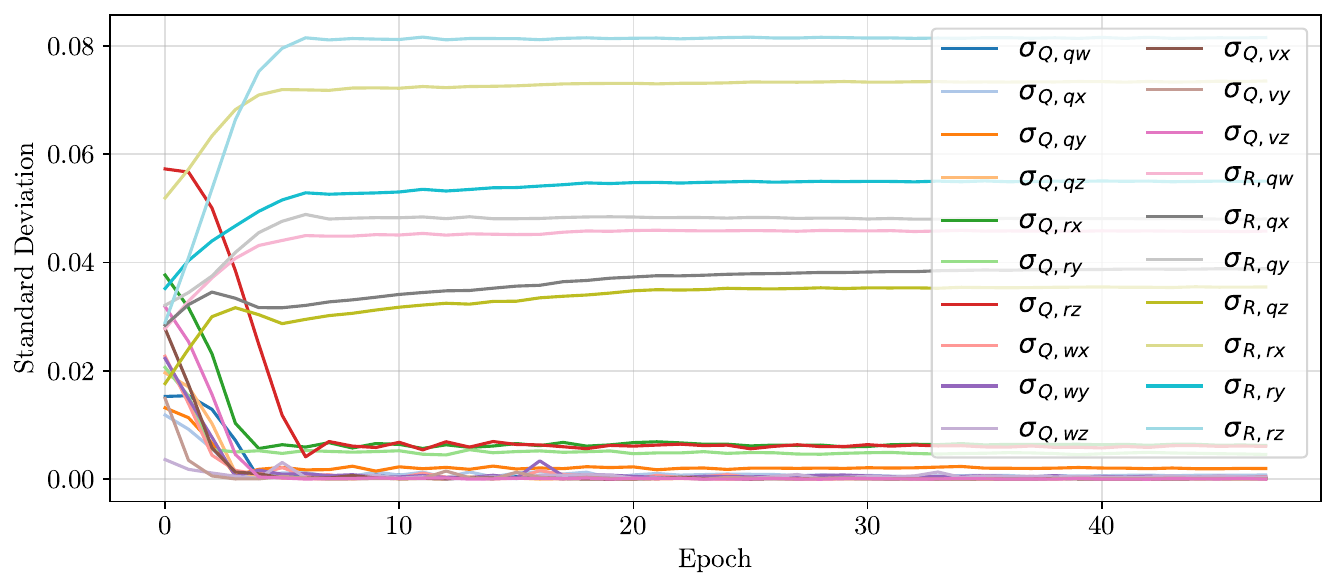}
\caption{Convergence behavior of the learned parameters during training with DS2. The $20$ noise covariance terms that the DFCNN is learning from the DS2 are tracked over each epoch and depicted in this graph.}
\label{fig:training_sigmas_ds2}
\end{figure}

To assess the learned parameter convergence, we consider the evolution of the learned noise covariance parameters in Fig.~\ref{fig:training_sigmas_ds2}. 
Similar to training convergence, the learned noise covariance parameters start converging after approximately six epochs. 
After that, there are only minor changes in the parameters, and at epoch $27$, the parameters remain relatively stable, yielding the lowest loss.

\begin{table}[!tb]
\centering
\caption{Final results of the training sessions with the best found hyperparameters for each dataset}
\begin{tabular}{lllll}
\hline
\hline
\addlinespace
Dataset & Best val. & Best val. &      Test & Computation time \\
        &     epoch & RMSE loss & RMSE loss & $[\si{\second}/\mathrm{epoch}]$ \\
\addlinespace
\toprule
\addlinespace
DS1 & $49$ & $0.0059$ & $0.0067$ & $39.4286$ \\
DS2 & $27$ & $0.0176$ & $0.0189$ & $41.2500$ \\
\addlinespace
\hline
\hline
\end{tabular}
\label{tab:training_results}
\end{table}

Table~\ref{tab:training_results} summarizes the final training results for both datasets, including the best validation epoch, best validation RMSE, test RMSE, and computation time per epoch. 
Notably, the losses for both datasets are fairly small, indicating high EKF accuracy. 
Additionally, the close match between validation and test losses suggests good generalization to unseen data. 
DS1 has significantly smaller losses compared to DS2, which can be attributed to the differences in angular velocities between the datasets. 
DS2's higher angular velocities lead to more significant and faster changes, making it more challenging to compensate for, and resulting in higher losses. Furthermore, the performance of the controller actuating the rotational motion of the target may also contribute to these differences, potentially introducing errors that manifest as inaccuracies in the ground truth. 

The computation times don't provide significant insights. 
While using GPUs for parallel computing might speed up convergence, the bottleneck may be the EKF state estimation, which cannot be parallelized efficiently and may take even longer when deployed on a GPU core instead of a computational faster CPU core. 
The best validation epochs indicate favorable and relatively fast convergence, considering the dimension of the parameter space being learned. 
Finally, Table~\ref{tab:training_results} can serve as a reference for tuning FlexKalmanNet under different settings, evaluating its performance with various KFs like the UKF, and comparing it to alternative networks like EKFNet or KalmanNet on the same dataset.

\subsection{Evaluation of the Learned Parameters} 
The learned parameters from DS1 and DS2 are summarized in Table~\ref{tab:parameters}. 
Notably, higher values for process noise covariances are learned from DS2, making the EKF more sensitive, which is attributed to the higher dynamics present in DS2. 
\begin{table}[!tb]
\centering
\caption{Learned standard deviation parameters from DS1 and DS2}
\begin{tabular}{l|ll}
\hline
\hline
\addlinespace
  Trained & Learned value & Learned value \\
parameter &       for DS1 &        or DS2 \\
\addlinespace
\toprule
\addlinespace
$\sigma_{R_{q_w}}$ & $0.042746$ & $0.045785$ \\
$\sigma_{R_{q_x}}$ & $0.036301$ & $0.038067$ \\
$\sigma_{R_{q_y}}$ & $0.048697$ & $0.048149$ \\
$\sigma_{R_{q_z}}$ & $0.029550$ & $0.035209$ \\[4pt]
$\sigma_{R_{r_x}}$ & $0.078810$ & $0.073300$ \\
$\sigma_{R_{r_y}}$ & $0.058190$ & $0.054914$ \\
$\sigma_{R_{r_z}}$ & $0.082663$ & $0.081457$ \\[4pt]
$\sigma_{Q_{q_w}}$ & $0.000161$ & $0.000145$ \\
$\sigma_{Q_{q_x}}$ & $0.000090$ & $0.000667$ \\
$\sigma_{Q_{q_y}}$ & $0.000071$ & $0.002006$ \\
$\sigma_{Q_{q_z}}$ & $0.000026$ & $0.000026$ \\[4pt]
$\sigma_{Q_{r_x}}$ & $0.000150$ & $0.006237$ \\
$\sigma_{Q_{r_y}}$ & $0.000010$ & $0.004650$ \\
$\sigma_{Q_{r_z}}$ & $0.000132$ & $0.006365$ \\[4pt]
$\sigma_{Q_{\omega_x}}$ & $0.000079$ & $0.000575$ \\
$\sigma_{Q_{\omega_y}}$ & $0.000145$ & $0.000515$ \\
$\sigma_{Q_{\omega_z}}$ & $0.000037$ & $0.000876$ \\[4pt]
$\sigma_{Q_{v_x}}$ & $0.000023$ & $0.000087$ \\
$\sigma_{Q_{v_y}}$ & $0.000028$ & $0.000022$ \\
$\sigma_{Q_{v_z}}$ & $0.000025$ & $0.000040$ \\
\addlinespace
\hline
\hline
\end{tabular}
\label{tab:parameters}
\end{table}

\noindent Striking a balance in parameter values is crucial; excessively high values can lead to noise sensitivity and hence overfitting, while too low values may result in underfitting~\cite{saha2013robustness}. 
The parameters for measurement noise covariances are very similar for both datasets, indicating that the same underlying noise can be learned from both datasets. 
However, the DFCNN adjusts the process noise covariance parameters to best-fit the present dynamics in each dataset during training, resulting in different sensitivities of the trained parameters.

\begin{figure}[!tb]
\centering
\includegraphics[width=0.485\textwidth]{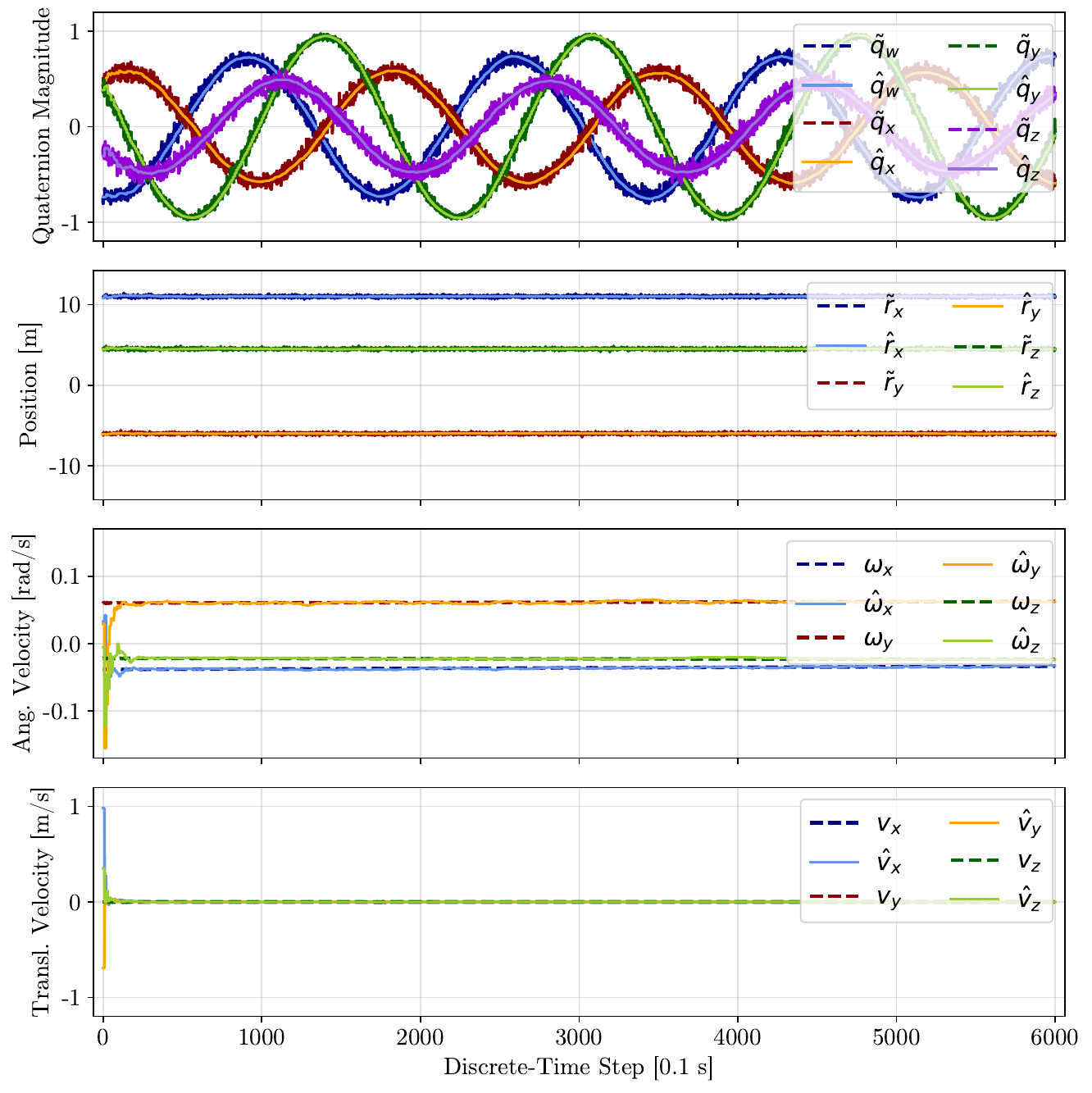}
\caption{EKF results using the trained parameters for DS1 applied on DS1. Each of the four subplots depicts one of the four motion components from the state vector: orientation, position, angular velocity and translational velocity. In every subplot, the EKF estimate and the corresponding measurement or ground truth are plotted, denoted as $\hat{x}$, $\tilde{x}$ and $x$, respectively.}
\label{fig:ekf_states_offline_ds1_ds1}
\end{figure}

Subsequent tests primarily focus on the trained parameters obtained from the more realistic dataset DS1. 
In Fig.~\ref{fig:ekf_states_offline_ds1_ds1}, an EKF with the learned parameters for DS1 is applied to the corresponding dataset (DS1) with moderate angular velocities. 
This initial test demonstrates several important aspects. 
Firstly, the EKF converges and remains stable over time. 
Secondly, all variables appear to be filtered effectively, with only small uncertainties observed around the ground truth of angular velocities. 
The EKF successfully tracks the quaternion evolution and follows the slightly changing angular velocities. 
The position and translational velocities, which lack significant changes, are correctly estimated by the EKF. 
Moving on to DS2, please note that the related plots have been omitted for the sake of brevity. 
Trained parameters obtained from DS2 and applied to the corresponding dataset DS2 exhibit similar behavior, resulting in accurate estimations. 
However, when we apply the learned parameters from DS1 to DS2, the estimates remain accurate but exhibit a slight lag behind the true states. 
This lag can be attributed to the lack of learned sensitivity to the higher dynamics present in DS2. Conversely, when applying learned parameters from DS2 to DS1, the estimates tend to overreact to noise inputs due to the higher values learned for the noise covariances. 

\begin{figure}[!tb]
\centering
\includegraphics[width=0.485\textwidth]{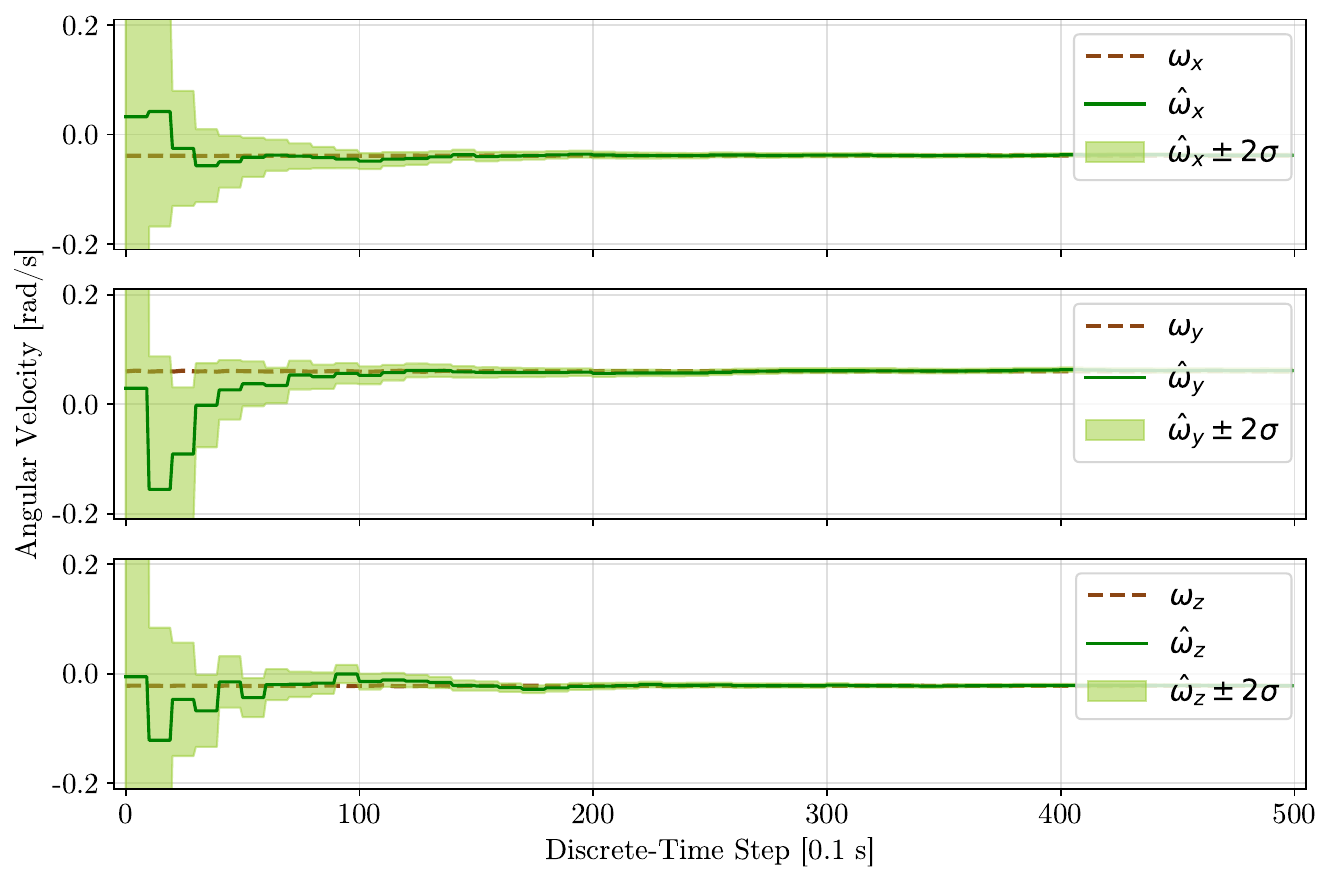}
\caption{Uncertainties of the EKF angular velocity estimates using the trained parameters from DS1 on DS1. Each of the three subplots depicts one of the components of the angular velocities. In every subplot, the EKF estimate and the corresponding ground truth, denoted as $\hat{x}$ and $x$, and the uncertainty space are plotted. For the uncertainty space, the double standard deviation ($2 \sigma$) of the corresponding error state feature is utilized.}
\label{fig:ekf_uncertainties_angular_velocities_ds1}
\end{figure}

To evaluate the accuracy and stability of the filter, we examine the standard deviations of each feature of the EKF error state. 
We conduct the EKF runs on the first $500$ samples of DS1 to enhance the resolution of the settling phase. 
As representative feature, the evolution of uncertainties in angular velocities during the EKF filtering is depicted in Fig.~\ref{fig:ekf_uncertainties_angular_velocities_ds1}.
The uncertainty is calculated as a double standard deviation, representing the space within which the estimate will fall with a $95.4\,\%$ probability. 
When observing the uncertainties of angular velocities, we notice a significant decrease within the first $100$ to $150$ discrete-time steps. 
After this phase, the uncertainty boundaries start converging towards the ground truth. 
This plot also facilitates the analysis of settling time, which correlates with the phase of a significant decrease in state uncertainty. 

\begin{figure}[!tb]
\centering
\includegraphics[width=0.485\textwidth]{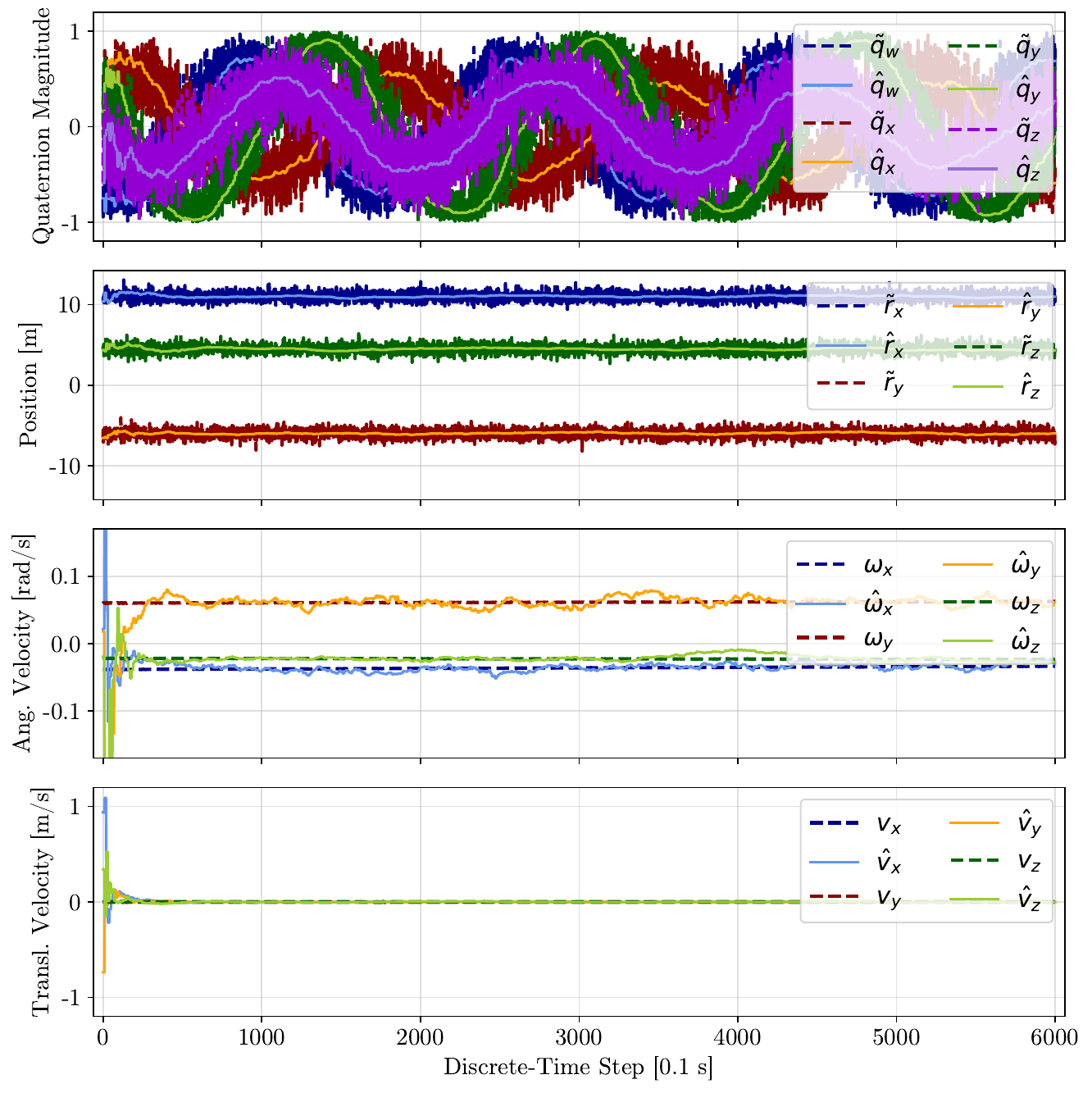}
\caption{EKF results for the trained DS1 parameters on DS1 with five times larger noise magnitude. Each of the four subplots depicts one of the four motion components from the state vector. In every subplot, the EKF estimate and the corresponding measurement or ground truth are plotted, denoted as $\hat{x}$, $\tilde{x}$ and $x$, respectively.}
\label{fig:ekf_states_noise_5_ds1}
\end{figure}

To assess robustness to noise, the learned parameters from DS1 are tested on DS1 with increased noise, five times larger than during training $(\sigma_R = 0.5)$. 
Fig.~\ref{fig:ekf_states_noise_5_ds1} depicts the results of the EKF run with DS1 parameters on DS1, applying the increased noise in the data. 
The impact of large noise is visible in the ground truth of the orientation and position subplots. 
Despite the noise magnitude being five times larger, the translational motion estimates remain stable. 
The rotational components exhibit uncertainties in angular velocity estimates, although they are still within a relatively small range, indicating accurate estimates despite the increased noise. 

\begin{figure}[!tb]
\centering
\includegraphics[width=0.485\textwidth]{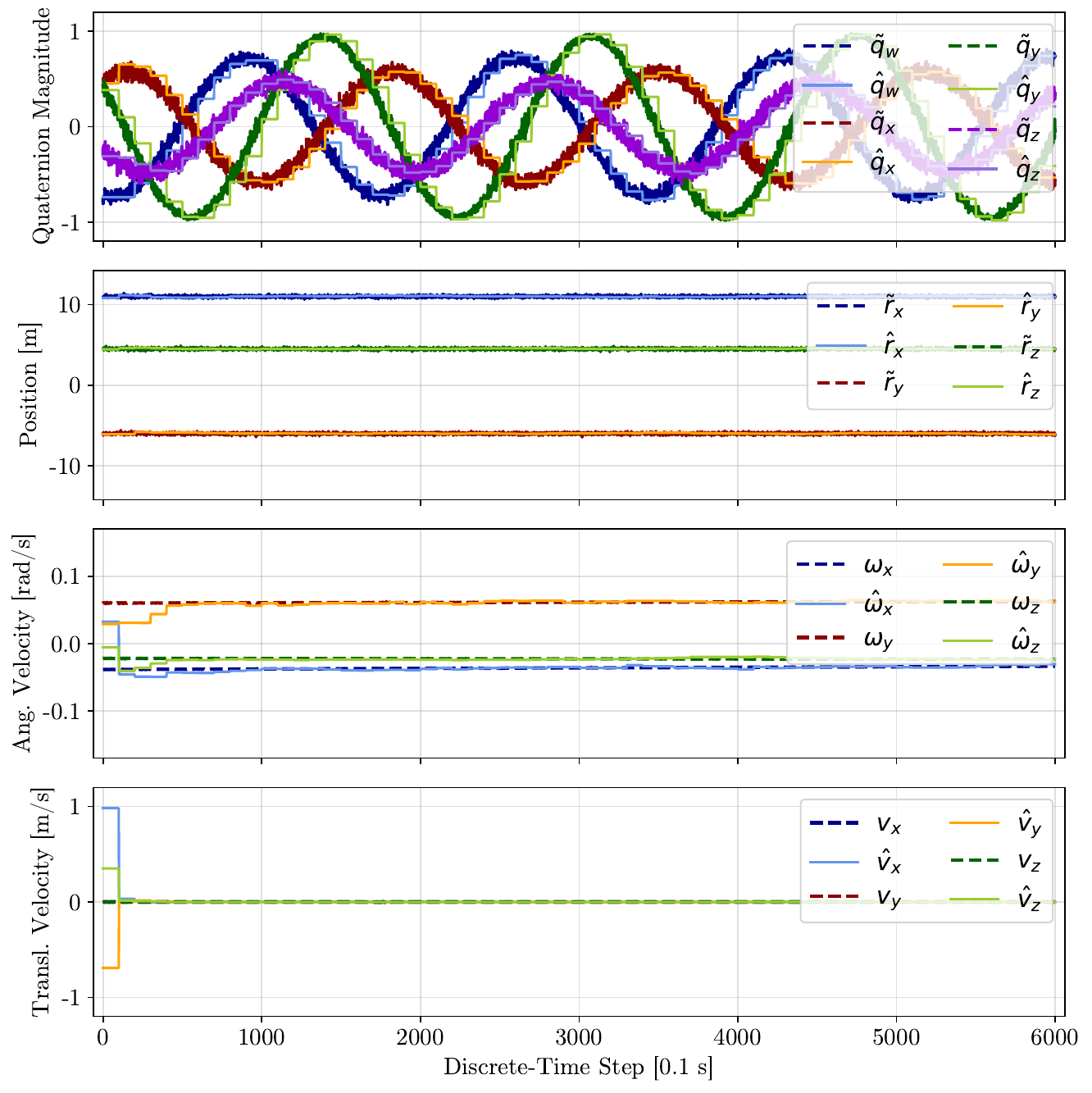}
\caption{EKF results for the trained DS1 parameters on DS1 with increased lack of data. Each of the four subplots depicts one of the four motion components from the state vector. In every subplot, the EKF estimate and the corresponding measurement or ground truth are plotted, denoted as $\hat{x}$, $\tilde{x}$ and $x$, respectively. The square-like shapes in the data are caused by an introduced lack of data of a discrete-time step size of $100$.}
\label{fig:ekf_states_steps_100_ds1}
\end{figure}

To assess robustness in scenarios with limited  data, we extend the number of discrete-time steps without measurements from $10$ to $100$. 
This simulates scenarios with low measurement update rates or potential data dropout. 
Fig.~\ref{fig:ekf_states_steps_100_ds1} presents the results of this stability test using the trained parameters from DS1 on DS1.
The settling time appears to increase, but the filter remains stable over time. 
The simulations of increased data dropout reaffirm the robustness of the learned parameters.

Finally, the accuracy of EKF runs using the parameters trained with DS1 and the manually tuned parameters are compared using the RMSE metric. 
The results are presented in Table~\ref{tab:manual}. 
\begin{table}[!tb]
\centering
\caption{RMSE of each variable of the state vector for EKF runs using learned parameters from DS1 and using the manually tuned parameters, both run on DS1}
\begin{tabular}{lll}
\hline
\hline
\addlinespace
   State & RMSE &          RMSE \\
variable &  DS1 & manual tuning \\
\addlinespace
\toprule
     $q_w$ & $0.044$ & $0.047$ \\
     $q_x$ & $0.046$ & $0.050$ \\
     $q_y$ & $0.039$ & $0.041$ \\
     $q_z$ & $0.047$ & $0.052$ \\
     $r_x$ & $0.102$ & $0.103$ \\
     $r_y$ & $0.103$ & $0.103$ \\
     $r_z$ & $0.101$ & $0.103$ \\
$\omega_x$ & $0.001$ & $0.008$ \\
$\omega_y$ & $0.001$ & $0.008$ \\
$\omega_z$ & $0.001$ & $0.008$ \\
     $v_x$ & $0.002$ & $0.002$ \\
     $v_y$ & $0.001$ & $0.001$ \\
     $v_z$ & $0.001$ & $0.001$ \\
\addlinespace
\hline
\hline
\end{tabular}
\label{tab:manual}
\end{table}
It shows that the RMSE of all features estimated by the EKF using parameters from DS1 is less than or equal to those using manually tuned parameters. 
The rotational components, which are more challenging to tune due to faster dynamics, are all more accurate when using the trained parameters. 
Translational features are similar, indicating that the initial estimate for manually tuned parameters of translational motion is well-selected. 
A reduction in process covariance magnitudes may lead to less sensitivity to noise and better results for manually tuned parameters. 
This manual tuning process is precisely the operation that FlexKalmanNet aims to replace, as it involves the true standard deviation of measurement noise, which is often unavailable, making FlexKalmanNet essential to find it.

Moreover, Table~\ref{tab:manual} demonstrates that all state features estimated using parameters from DS1 are very accurate. 
For example, the RMSE of $\omega_x$ is $\SI{0.001}{\radian}/\si{\second}$, equivalent to $\SI{0.057}{\degree}/\si{\second}$. 
The overview of feature-wise errors also shows that the features introduced as noisy measurements exhibit significantly larger RMSE compared to the twist features. 
This difference can be attributed to the fact that, unlike orientations and positions, the time derivatives (angular and translational velocities) of these features are not directly measured and are more sensitive to inaccuracies in the underlying motion model.

\section{Discussion and Outlook}\label{sec:conclusion_and_outlook}
Motion estimation algorithms are essential in various fields, including space operations, where precise parameter tuning can be challenging. 
This work focuses on developing an AI-based algorithm for motion estimation in the context of tumbling objects, using the Astrobee free flyers as an example. 
The state of interest encompasses pose and twist, and data is generated through a ROS-based simulation environment for the ARS.
To address this challenge, a framework called FlexKalmanNet is proposed. FlexKalmanNet combines KF variants with a DFCNN. 
This framework is promising because it can potentially capture complex and nonlinear patterns more effectively than RNN-based architectures. 
FlexKalmanNet, with its modular structure, allows for the learning of any parameter of different KF variants. 

In this work, an EKF is incorporated into FlexKalmanNet to estimate the states of nonlinear systems efficiently. 
The states to be estimated include the pose and twist of a target. Noisy measurement comprise position and orientation data, represented as unit quaternions.   
The framework demonstrates that learning the unknown parameters of the EKF, specifically the covariances of the process $\boldsymbol{Q}$ and measurement noise $\boldsymbol{R}$, is feasible using the DFCNN architecture. 
The testing of the EKF with the trained parameters reveals several key findings: It remains stable over time, is robust against larger noise levels than those present in the training dataset, and is also robust against measurement lags. 
EKF estimates are accurate but depends on the sensitivity learned through the data; higher sensitivity improves reaction to state changes but may decrease accuracy. 
The settling of the EKF takes approximately $10$ to $15$ seconds to converge, which is acceptable for spacecraft rendezvous scenarios. 
After convergence, the trained EKF provides reliably accurate filtered data. 
To reduce the settling phase, introducing chaser motion can increase certainty in target state estimates. 

Comparing the EKF results using the learned parameters for the most realistic dataset (DS1) to manually tuned parameters, the EKF with learned parameters outperforms the manually tuned one. 
Manual tuning requires prior knowledge of noise characteristics, while FlexKalmanNet can learn parameters without prior information. 
However, FlexKalmanNet has two limitations: it must be trained before deployment and requires target data and corresponding ground truth, which may not always be available.

Future evaluations of FlexKalmanNet may involve comparisons with state-of-the-art approaches like EKFNET or KalmanNet, exploring the use of UKF, and adopting more advanced dynamics models. To improve accuracy, investigating learning beyond just the diagonal covariance matrix entries and considering potential noise feature dependencies is a possibility. Additionally, using larger batch sizes for training to capture more measurement data context could be explored.

\bibliography{literature/IEEEabrv,literature/paper}

\end{document}